\documentclass[10pt,twocolumn,letterpaper]{article}
\usepackage{iccv}              

\usepackage{multirow}
\usepackage{colortbl}
\usepackage{booktabs}
\usepackage{amssymb}
\usepackage{bbding}
\usepackage{pifont}
\usepackage{url} 

\definecolor{iccvblue}{rgb}{0.21,0.49,0.74}
\usepackage[pagebackref,breaklinks,colorlinks,allcolors=iccvblue]{hyperref}

\def\paperID{8076} 
\def\confName{ICCV}
\def\confYear{2025}

\title{CVFusion: Cross-View Fusion of 4D Radar and Camera for 3D Object Detection}

\author{Hanzhi Zhong \quad Zhiyu Xiang$^*$ \quad Ruoyu Xu \quad Jingyun Fu \quad Peng Xu \\ Shaohong Wang \quad Zhihao Yang \quad Tianyu Pu \quad Eryun Liu\\
{\normalsize Zhejiang University}\\
\href{https://github.com/zhzhzhzhzhz/CVFusion}{https://github.com/zhzhzhzhzhz/CVFusion}}

\begin{document}
\maketitle
\begin{abstract}
4D radar has received significant attention in autonomous driving thanks to its robustness under adverse weathers. Due to the sparse points and noisy measurements of the 4D radar, most of the research finish the 3D object detection task by integrating images from camera and perform modality fusion in BEV space. However, the potential of the radar and the fusion mechanism is still largely unexplored, hindering the performance improvement. In this study, we propose a cross-view two-stage fusion network called CVFusion. In the first stage, we design a radar guided iterative (RGIter) BEV fusion module to generate high-recall 3D proposal boxes. In the second stage, we aggregate features from multiple heterogeneous views including points, image, and BEV for each proposal. These comprehensive instance level features greatly help refine the proposals and generate high-quality predictions. Extensive experiments on public datasets show that our method outperforms the previous state-of-the-art methods by a large margin, with 9.10\% and 3.68\% mAP improvements on View-of-Delft (VoD) and TJ4DRadSet, respectively. Our code will be made publicly available.
\end{abstract}    
\section{Introduction}
\label{sec:intro}

3D object detection is a vital perception task for autonomous vehicles. In recent years 4D millimeter-wave radar has received significant attention thanks to its robustness under adverse environments, such as rainy, foggy and snowy weathers. Comparing with the traditional 3D radar which only provides azimuthal, range, and Doppler velocity measurements, 4D radar is capable of acquiring an extra valuable elevation measurement, producing similar 3D point clouds like LiDAR. However, the point cloud of 4D radar is highly sparse compared to the LiDAR, with point number of 0.1k$\sim$2k/frame versus 10k$\sim $100k/frame. Moreover, radar's measurements are much noisier than that of LiDAR due to the multipath effect of the millimeter wave. Therefore, relying solely on 4D radar for 3D object detection is less satisfactory. Most of the current research focus on fusing the 4D radar with images from a camera to improve the detection performance.

\begin{figure}[t]
\centering
\includegraphics[width=0.45\textwidth]{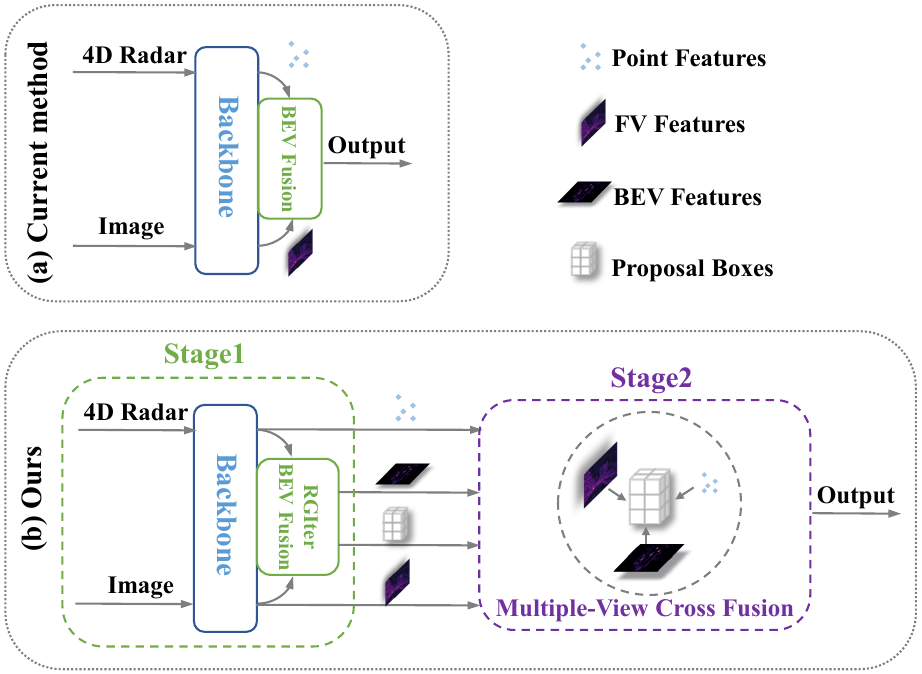}
\caption{Comparison between the current method and our two-stage pipeline. We introduce stage2 to aggregate multiple-view features and refine the proposals. In the stage1, we introduce radar guided iterative (RGIter) BEV fusion to produce high quality proposals.}
\label{mainfig}
\end{figure}

Current 4D radar and camera fusion works \cite{rcfusion,lxl} adopt one-stage pipeline and transform the features of the two modalities separately into a unified BEV space for fusion, as shown in \cref{mainfig}(a). Although compact and effective, this pipeline suffers from the feature misalignment problem. Due to the depth ambiguity of the monocular images, position drift happens when transforming the front-view (FV) feature of the image to the BEV view. Meanwhile, unlike LiDAR, the sparse and noisy radar points are not able to provide sufficient and precise 3D information for the images. Therefore, directly concatenating BEV features of the radar and camera will inevitably cause misalignments, especially for small objects, resulting in position drift of the predictions and false positives.

To alleviate this problem, we propose a two-stage-based radar-camera fusion network.  Instead of generating final results at once like the one-stage network, our model first generates some proposal boxes and then extracts features within the proposals for refinement. In this way, the entire task is shared by the two stages and the feature alignment could be gradually refined by different fusion mechanism. To improve the recall of the proposals in stage1, we introduce a Radar-Guided Iterative (RGIter) BEV fusion module which can achieve better alignments of features. In the second stage, we implement multi-view cross-modality fusion to refine the proposals. The difference between the current methods and ours are shown in \cref{mainfig}.

In order to fully explore the characteristics of different views from the heterogeneous modalities, we design two branches ,i.e., Point-Guided Fusion (PGF) and Grid-Guided Fusion (GGF) for comprehensive feature aggregation in the stage2. The PGF is responsible for retrieving the radar point and fusing the image front-view (FV) features within the proposals. Considering some proposal boxes may not be assigned with valuable features due to the lack of radar points, we further design GGF which can assign and fuse the features according to the spatial grids within the proposals. It aggregates image FV feature and the hybrid-BEV feature in a sequential way. Thanks to the spatial orthogonal nature of the FV and BEV, the bounding box of the proposal could be further refined. Our method has been extensively tested on the public VoD and TJ4DRadSet dataset, and achieved significant performance improvements compared to the state-of-art works. 

To summarize, our contributions are as follows:
\begin{itemize}
\item We propose a novel cross-view fusion method for 3D object detection based on 4D radar and camera. It features fulfilling the heterogeneous multi-view fusion in a gradual two-stage way. To the best of our knowledge, it is the first model that fusing 4D radar and camera in a two-stage pipeline;
\item A radar guided iterative BEV fusion module is proposed in stage1 to better align the BEV features and generate high-quality proposals;
\item Point- and grid-guided fusion modules are proposed in the stage2 to effectively fuse multiple-view features;
\item We achieve the new state-of-the-art performance on VoD and TJ4DRadSet. Extensive experiments demonstrate the effectiveness of the critical components of CVFusion.
\end{itemize}

\section{Related Work}
\label{sec:related_work}

\subsection{4D Radar-based 3D Object Detection}

The measurement of 4D radar can be categorized into two types: radar tensor and point cloud \cite{yao2024exploringradardatarepresentations}. 4D Radar Tensors (4DRT) are usually in the form of continuous power measurements along the Doppler, range, azimuth, and elevation dimensions \cite{Rebut_2022_CVPR, paek2022kradar}. RTNH \cite{paek2022kradar} and RTNH+ \cite{kong2023rtnh+} are 4DRT-based 3D object detection methods. After applying the constant-false-alarm-rate (CFAR) algorithm on 4DRT, popular Radar Point Cloud (RPC) can be obtained, with each point equipped with xyz, Doppler velocity, and radar cross-section (RCS). Most of the existing methods are based on the form of point cloud for its generality. Although LiDAR-based 3D object detection networks \cite{pointnet, lang2019pointpillars, second} can be directly applied for 4D radar points, there are some works specifically designed for 4D radar. RPFA-Net \cite{rpfa} improves the accuracy of heading angle estimation by introducing a self-attention layer for the pillar-feature. RadarMFNet \cite{radarMFNet} aggregates multiple frames to obtain denser radar points. MVFAN \cite{yan2023mvfan} proposes a multi-feature based network, explicitly exploiting the valuable Doppler velocity and reflectivity of 4D radar measurements. SMURF \cite{liu2023smurf} introduces an additional density feature from kernel density estimation (KDE) for data denoising. SMIFormer \cite{shi2023smiformer} learns 3D spatial features by multi-view interaction with transformer. 

\subsection{Radar-Camera Fusion for 3D Object Detection}
According to the type and measurement form of radar, the radar-camera fusion methods can be divided into three categories: 3D radar-camera fusion, 4DRT-camera fusion, and 4D RPC-camera fusion. Early 3D-radar and camera fusion methods \cite{nabati2021centerfusion, wu2023mvfusion} employ front-view (FV) fusion which projects radar points onto the image and integrating them as supplementary features  before bounding box regression. Later approaches \cite{simplebev, RCBEV} transform both the camera and radar features into Bird’s-Eye-View (BEV) for fusion. CRN \cite{crn} aggregates camera and radar features on BEV through a multimodal deformable attention.  EchoFusion \cite{liu2023echoes} and DPFT \cite{fent2024dpft} belong to the methods of 4DRT-camera fusion. EchoFusion proposes a Polar-Aligned Attention module to guide the network to learn effective radar and image features. DPFT proposes an efficient dual perspective fusion approach to simplify the fusion pipeline. For 4D radar point cloud based fusion, RCFusion \cite{rcfusion} employs an interactive attention module to adaptively fuse radar and image feature on BEV. LXL \cite{lxl} proposes a radar occupancy-assisted depth sampling net to improve the quality of image voxel features. To alleviate radar azimuth errors, RCBEVDet \cite{rcbevdet} uses a CAMF module to align the image-radar BEV feature dynamically.

\begin{figure*}[t]
\centering
\includegraphics[width=0.95\textwidth]
{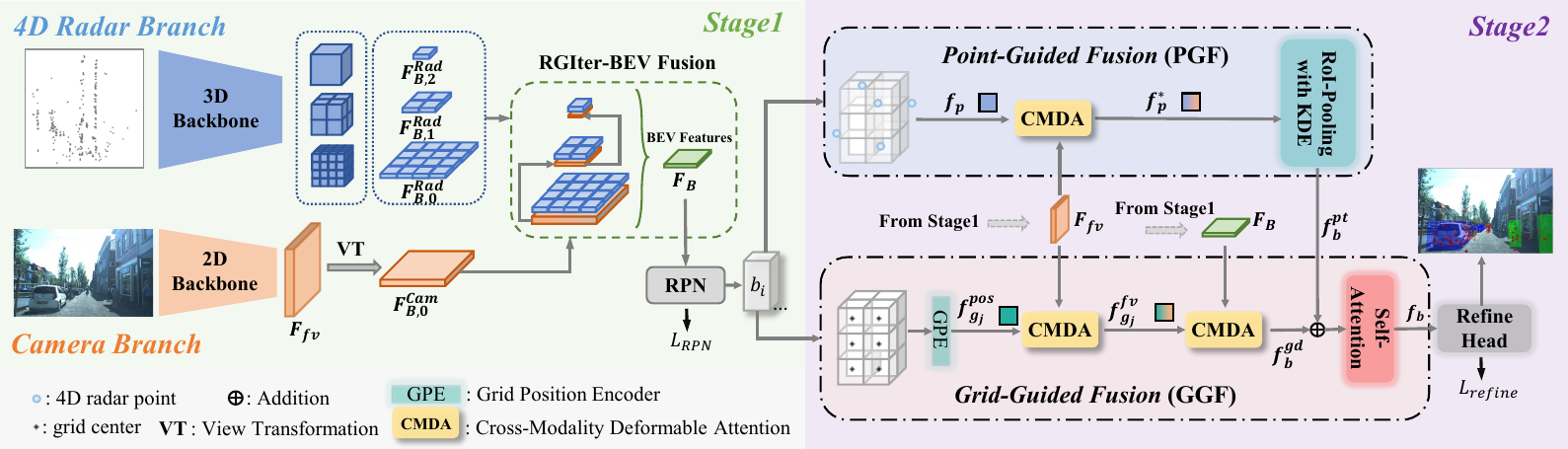}
\caption{Overall architecture of CVFusion. The first stage is responsible for preliminary feature fusion and proposal generation, while the second stage refines the proposals with delicate cross-view fusion.}
\label{architecture}
\end{figure*}

\subsection{Proposal-based 3D Object Detection}

Other than the one-stage scheme, proposal-based two stage pipeline is also popular in object detection. Among them, \cite{voxelrcnn,pvrcnn,pdv} are the typical LiDAR-only two-stage networks. As a LiDAR-camera fusion network, LoGoNet \cite{logonet} projects the proposal boxes onto the image plane and performs attention-based fusion. A few 3D radar-camera fusion methods also lie in this category. GRIF Net \cite{grif} projects 3D proposal boxes onto the image plane and obtains image features to correct the proposals. CRAFT \cite{craft} utilizes a Soft Polar Association module to associates radar points with the object proposals. Based on the DETR \cite{detr} paradigm, TransCAR \cite{transcar} utilizes vision-updated 3D object query to interact with radar features. 

Our CVFusion is the first 4D radar-camera fusion network adopting the two-stage paradigm. Compared to the current one-stage methods \cite{rcfusion, lxl}, we generate proposal boxes through radar guided iterative BEV fusion and refine the proposals in the second stage. Comparing with the LiDAR-based two-stage networks \cite{logonet}, we fully consider the high sparsity and noise that associated with the 4D radar point clouds, and delicately design the multi-view cross-modality feature fusion in the proposal-refining stage.

\section{Method}
\subsection{Overview}

The overall architecture of our CVFusion is shown in \cref{architecture}. In the Stage1, the 4D radar points and image features are first extracted by the 3D and 2D backbones, and transformed to the BEV plane, respectively. Then these BEV features are fed into the Radar Guided Iterative BEV (RGIter-BEV) fusion module for further processing. The resulting BEV fusion map is sent to the region proposal network (RPN) to generate proposal boxes. In the Stage2, for each proposal we aggregate cross-view features from point, image, and BEV by the Point-Guided Fusion (PGF) and Grid-Guided Fusion (GGF) modules, and generates final refined predictions.

\subsection{Stage1: Feature Extraction and BEV Fusion}

\subsubsection{4D Radar and Camera Feature Extraction}
For the 4D radar branch, we use SECOND \cite{second} as backbone to extract radar features. The input radar point clouds are denoted as $P = \{{(x_i, y_i, z_i, f_{i})\}}_{i=1}^N$, where $(x_i, y_i, z_i)$ is the spatial coordinate of $i$-th point, $f_{i} \in \mathbb{R}^{C_p}$ is the measurements containing the Doppler velocity and RCS, $N$ is the number of points in the cloud. These radar points will be voxelated and fed into the backbone network. After multi-layer 3D sparse convolution, multi-scale radar voxel features $F_{V,j}^{Rad} \in \mathbb{R}^{\frac{X}{2^j} \times \frac{Y}{2^j} \times \frac{Z}{2^j}\times C_{V_j}} , j=0,1,2,3,4$ are obtained, where $(\frac{X}{2^j}, \frac{Y}{2^j}, \frac{Z}{2^j})$ represents the size of voxels and the $C_{V,j}$ is the number of channels of the voxel feature at scale $j$. 
Then, we choose to convert voxel features with $j=2,3,4$ into BEV features, obtaining $F_{B,k}^{Rad} \in \mathbb{R}^{\frac{X}{4*2^k} \times \frac{Y}{4*2^k} \times C_{B_k}}$, where $k=0,1,2$ and $C_{B,k}$ is the number of channels of the BEV features.

For the camera branch, the input image $I$ is fed into a pre-trained 2D backbone \cite{liu2021Swin} to extract image features $ F_I \in \mathbb{R}^{ \frac{W_I}{4} \times \frac{H_I}{4} \times C_I}$, where $W_I$, $H_I$ are the width and height of the image, $C_I$ is the number of channels.
In the step of view transformation, inspired by \cite{reading2021categorical, lxl}, we adopt the depth-based view transformation strategy to convert the front view features $F_I$ into BEV features $F_{B}^{Cam} \in \mathbb{R}^{\frac{X}{4} \times \frac{Y}{4} \times C_{B}}$, where $C_{B}$ is the number of channels of the camera BEV feature. 

In particular, the spatial resolution of $F_{B}^{Cam}$ is kept the same as $F_{B,k=0}^{Rad}$ , which will facilitate the subsequent iterative BEV fusion.

\subsubsection{Radar Guided Iterative BEV Fusion}

Considering the depth uncertainty of the projected image BEV feature, we propose a radar guided iterative BEV fusion module to strengthen the position embedding of the image feature and improve the fused features thereby.

\begin{figure}[t]
\centering
\includegraphics[width=0.75\columnwidth]{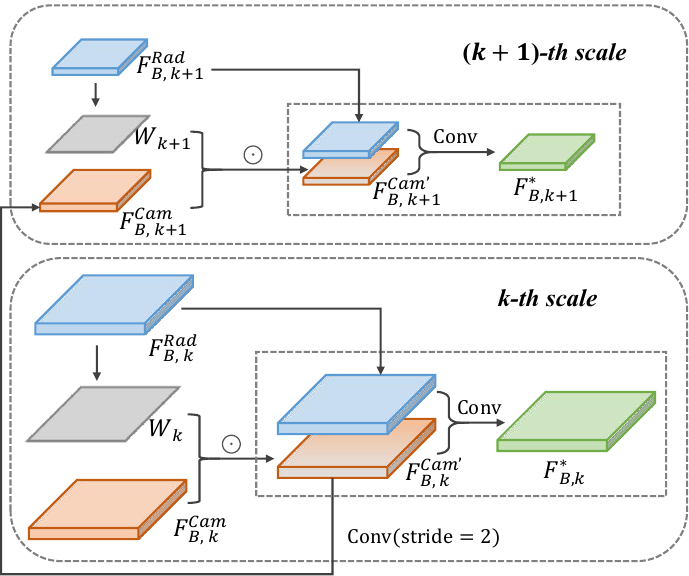} 
\caption{The structure of the RGIter-BEV Fusion module. The inputting image features at the $(k+1)$-th scale are iteratively generated by convolving and down-sampling the radar-occupancy-weighted image features at $k$-th scale.}
\label{RGMS fusion}
\end{figure}

We adopt an iterative mechanism to gradually refine the image BEV features for fusion. As shown in \cref{RGMS fusion}, at each scale $k$ we first generate a weight map which represents the occupancy probability of the BEV grid, from the corresponding radar BEV feature, with
\begin{align}
    \label{eq:weight map}
    W_{k} = \mathrm{Sigmoid}(\mathrm{Conv2d}(F_{B,k}^{Rad})), k=0,1,2
\end{align}

Then we multiply the $W_{k}$ to the corresponding camera BEV feature map to strengthen the position information of the camera BEV feature, as 
\begin{align}
    \label{eq:msw-conv}
    F_{B,k}^{Cam'} &= F_{B,k}^{Cam} \odot W_{k},
\end{align}
\noindent

where $\odot$ represents element-wise multiplication with broadcasting, $F_{B,k}^{Cam'}$ and $F_{B,k}^{Cam}$ are the updated and the original camera BEV feature at scale $k$. When $k=0$, $F_{B,k}^{Cam}=F_{B}^{Cam}$. Otherwise, $F_{B,k+1}^{Cam}$ is iteratively produced from the $k$-th scale with
\begin{align}
    \label{eq:ms-img}
    F_{B,k+1}^{Cam} &= \mathrm{Conv2d}(F_{B,k}^{Cam'}, \mathrm{stride}=2), k = 1,2
\end{align}
\noindent

where $\mathrm{Conv2d}$ and $\mathrm{stride}$ represent common 2D convolution and down-sampling with a stride. 

The radar feature maps $F_{B,k}^{Rad}$ and the image feature maps $F_{B,k}^{Cam'}$ of the same scale are then concatenated and convolved to obtain the fused BEV feature $F_{B,k}^{*}$ with 
\begin{equation}
\begin{array}{c}
F_{B,k}^{*} = \mathrm{Conv2d}([F_{B,k}^{Rad}, F_{B,k}^{Cam'}]), k=0,1,2
\end{array}
\label{eqn:final-concat}
\end{equation}

At last, $F_{B,k}^{*}$ at all three scales are unified to the same scale of $k=1$ to obtain the final BEV fusion feature $F_{B}$:
\begin{equation}
\begin{array}{c}
F_{B} = \mathrm{Conv2d}([\mathrm{Down}(F_{B,0}^{*}),F_{B,1}^{*},\mathrm{Up}(F_{B,2}^{*})]),
\end{array}
\label{eqn:3-concat}
\end{equation}

where, $[\ \cdot \ ]$ means concatenation of feature maps, and $\mathrm{Down}$ and $\mathrm{Up}$ represent reducing and increasing the resolution of feature maps by 2 through bilinear interpolation, respectively. The BEV fusion map $F_{B}$ is then fed into the region proposal network (RPN) network to generate 3D proposal boxes $B = \{b_1, b_2, ...,b_n\}$.

\subsection{Stage2: Multiple-View Cross-Modality Fusion}

The task of the second stage is to refine the objects’ proposals produced in the stage1. To fully explore the potential of different sensors, a delicate multi-view-based cross-modality fusion module is proposed. As shown in \cref{architecture}, the module is composed of two branches, i.e., Point-Guided Fusion (PGF) and Grid-Guided Fusion (GGF). 

\subsubsection{Point-Guided Fusion (PGF)}
\label{sec:Point Guided Fusion Module}

\begin{figure}[t]
\centering
\includegraphics[width=0.45\textwidth]{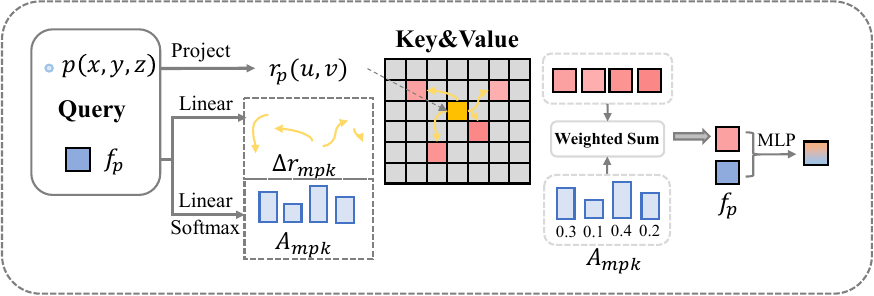}
\caption{The architecture of the Cross Modality Deformable Attention (CMDA) block. The position of query points are projected onto the image plane and the resulting queried features are obtained by the deformable attention. The querying results are concatenated with the original ones and then fused by an MLP layer.}
\label{stage2}
\end{figure}

Radar point clouds have strong spatial prior information, while images have dense semantic information. The PGF module aims to fuse radar and image features in the form of points. It is accomplished by a cross-modality deformable attention (CMDA) block, as shown in \cref{stage2}. Given the radar points within the proposal, their $(x,y,z)$  coordinates and the corresponding voxel features $F_{V,i=2}^{Rad}$ are retrieved from stage1. For each point $\boldsymbol{p}$ with its voxel feature $f_p$, we first obtain the projected point $\boldsymbol{r_p}=(u, v)$ in the image plane as:
\begin{align}
    \label{eq:project}
    {\boldsymbol{r_p}(u, v)}^T &= \mathbf{T} \cdot {\boldsymbol{p}(x, y, z)}^T
\end{align}
\noindent
where $(u, v)$ is the projected pixel location on the image plane, $\mathbf{T}$ is the projection matrix of the camera.

Following \cite{deformable}, we use the point feature ${f_{p}}$ as the \textit{Query}, and the image feature $F_I$ as the \textit{Key} and \textit{Value} in the deformable-attention. The queried image feature ${f}_I^{*}$ can then be obtained as:
\begin{equation}
\begin{array}{c}
{f_{I}^{*}} = \mathrm{CMDA}(f_{p},{r}_p, F_{I})= \\ 
\sum\limits_{m = 1}^M {{W_m}} \left[ \sum\limits_{k = 1}^K {{A_{mpk}}\cdot(W_m^{'}{F_{I}({{r}}_p} + \Delta {{{r}}_{mpk})})} \right],\\
\end{array}
\label{eqn:pgf}
\end{equation}
where ${A_{mpk}} = \mathrm{Softmax}(\mathrm{Linear}(f_{p}))$ and $\Delta {{{r}}_{mpk}} = \mathrm{Linear}(f_{p})$ denote the attention weight and sampling offset of the $k$-th sampling point in the $m$-th attention head, respectively. $W_m$ and $W_m^{'}$ are the learnable weights, $M$ is the number of self-attention heads and $K$ is the number of the sampled points.

Then we concatenate the queried image features ${f_{I}^{*}}$ with the point features $f_{p}$, and fuse them through an MLP layer to obtain the image enhanced point features $f_{p}^{*}$:
\begin{align}
    \label{eq:pg_fusion}
    f_{p}^{*} = \mathrm{MLP}([f_{p},f_{I}^{*}]),
\end{align}
\noindent

The features $f_{p}^{*}$ are then fed into a grid-based RoI-Pooling layer for aggregation. The proposal boxes are divided into $U * U * U$ uniform grids, each of which aggregates nearby point features $f_p^*$ within a certain ball radii. To distinguish the isolated (possibly noisy) points from others, we further introduce KDE \cite{pdv} for each point and concatenate it into $f_p^*$. The KDE will produce higher values for points with denser neighbors in the ball. We use maxpooling for all the points within the grid. Finally, all grid features are realigned into a feature $f_b^{pt} \in \mathbb{R}^{U^3 \times C_{b}}$, where $b$ and $C_{b}$ represent the box and the number of channels, respectively.

\subsubsection{Grid-Guided Fusion (GGF)}

Considering the sparsity of 4D radar point clouds, a large number of proposals may contain few or even no radar points. Solely relying on PGF cannot guarantee valuable features for all proposal boxes, which will result in inferior detecting performance. We propose Grid-Guided Fusion to mitigate this problem.

In this module, we divide each proposal box into $U^3$ grids $\{g_1, \dots  ,g_{U^3}\}$ similar to RoI-Pooling and generate the features in the form of grid, regardless of the presence of radar points. Following \cite{logonet}, these grids are first encoded by the Grid Position Encoder (GPE) to generate initial grid features $f_{g_j}^{pos}$ as:
\begin{equation}
    f_{g_j}^{pos} = \mathrm{GPE}(g_j) = \mathrm{MLP}(\delta_{j}, c_b, \log(\left|N_{g_j} \right|+\epsilon)),
\end{equation}

where $\delta_{j} = g_j-c_B$ is the relative position of each grid from the proposal box center $c_b$, $|N_{g_j}|$ is the number of points in each grid $g_j$, and $\epsilon$ is a constant offset. 

As shown in \cref{architecture}, the initial grid features $f_{g_j}^{pos}$ will act as a \textit{Query} and be fused with the image front-view (FV) and the BEV view features sequentially by two CMDA blocks. The process of the first part is similar to the PGF module, except that the query positions are generated by the regular grid centers. The output of the first CMDA is the queried FV features $f_{g_j}^{fv}$ for each grid ${g_j}$.

With $f_{g_j}^{fv}$ at hand, we continue to fuse it with BEV features $F_B$ from Stage1, which includes fused comprehensive BEV features. With the FV features $f_{g_j}^{fv}$ as \textit{Query}, and the BEV features $F_B$ as \textit{Key} and \textit{Value}, the fused features are computed by the second CMDA block. In this process, the projected position ${r^B_j}(u^B,v^B)$ of the proposal grid $g_j(x,y)$ are obtained by:
\begin{equation}
\begin{array}{c}
\begin{bmatrix}
 u^B\\  v^B
\end{bmatrix}
=(
\begin{bmatrix}
 g_j(x)\\  g_j(y)
\end{bmatrix}-
\begin{bmatrix}
 X_{min}\\  Y_{min}
\end{bmatrix}
)^T \cdot 
\begin{bmatrix}
 {1}/{X_{size}}\\ {1}/{Y_{size}}
\end{bmatrix},
\end{array}
\label{eqn:bev_ref_points}
\end{equation}

where $X_{min}, Y_{min}$ are the minimum $x,y$ coordinates of the BEV map, $X_{size},Y_{size}$ are the size of the grids on $x$ and $y$ axis, respectively. 

After the second CMDA block, we obtain the FV-BEV joint features $f_{g_j}^{gd}$ for each grid $g_i$. A total of $U^3$ grid features $\{f_ {g_1}^{gd}, \dots  ,f_ {g_{U^3}}^{gd}\}$ in the proposal are then concatenated to obtain $f_b^{gd} \in \mathbb{R}^{U^3 \times C_{b}}$, where $C_{b}$ is the number of feature channels which is consistent with $f_b^{pt}$ in PGF.

Given the point-guided fusion feature $f_b^{pt}$ and the grid-guided fusion feature $f_b^{gd}$, we use a transformer-based self-attention network \cite{vaswani2017attention} to explore relationships between them and deeply fuse each other, as:
\begin{equation}
\begin{array}{c}
    f_b = \mathrm{SelfAttn}(f_b^{pt}+f_b^{gd})
\end{array}
\label{eqn:SA}
\end{equation}

where $f_b$ represents the final feature of the proposal box $b$. The resulting $f_b$ will be sent to the refine head to predict confidence and box offsets for the object, as shown in \cref{architecture}.

\subsection{Training Losses}

Following the optimizer and loss function settings of \cite{pvrcnn}, we trained our CVFusion in an end-to-end manner. The overall training loss $L$ of the network is composed of the RPN loss $L_{RPN}$ and the refine loss $L_{refine}$:
\begin{equation}
\begin{array}{c}
L = L_{RPN} + L_{refine}
\end{array}
\label{eqn:loss}
\end{equation}

where the refine loss $L_{refine}$ includes the confidence loss and the regression loss between the proposal and the groudtruth boxes.

\section{Experiments}
\subsection{Datasets and Settings}

\begin{table*}[t]
\centering
\resizebox{0.95\textwidth}{!}{
\begin{tabular}{cc|ccc|c|ccc|c|c}
\hline
\multirow{2}{*}{Method} & \multirow{2}{*}{Modality} & \multicolumn{4}{c|}{Entire Annotated Area (\%)} & \multicolumn{4}{c|}{Driving Corridor Area (\%)} & \multirow{2}{*}{FPS}\\ \cline{3-10} 
 &  & Car & Ped. & Cyclist & mAP & Car & Ped. & Cyclist & mAP \\ \hline
PointPillars$^*$ \cite{lang2019pointpillars} & L & 61.11 & 49.77 & 72.58 & 61.16 & 90.89 & 54.63 & 94.29 & 79.93 & 43.7\\ 
\hline
MVFAN \cite{yan2023mvfan}  & R & 34.05 &  27.27 &  57.14 & 39.42 & 69.81 & 38.65 & 84.87 &  64.38 & 45.1 \\
PointPillars$^*$ \cite{lang2019pointpillars}  & L$\to$R & 41.95 &  38.09 & 67.01 & 49.02 & 70.56 & 48.41 & 87.13 & 68.70 & 45.5 \\
SECOND$^*$ \cite{second} & L$\to$R & 39.53  & 36.23  & 65.68 & 47.14 & 71.85 & 46.95 & 86.68 & 68.49 & 51.3 \\ 
PV-RCNN$^\ddagger$$^*$ \cite{pvrcnn} & L$\to$R & 40.90  & 41.89  & 73.61 & 52.13 & 72.11 & 51.78 & 85.89 & 69.93 & 22.6 \\ 
SMIFormer \cite{shi2023smiformer} & R & 39.53 & 41.88 & 64.91 & 48.77 & 77.04 & 53.40 & 82.95 & 71.13 & 16.4 \\ 
SMURF \cite{liu2023smurf} & R & 42.31 & 39.09 & 71.50 & 50.97 & 71.74 & 50.54 & 86.87 & 69.72 & $\backslash$ \\ 
\hline
FUTR3D$^*$ \cite{futr3d}  & (${R_{3D}}$$\to$R)+C & 46.01  & 35.11 & 65.98 &  49.03 & 78.66 & 43.10 & 86.19 & 69.32 & 7.3 \\
BEVFusion$^*$ \cite{liu2023bevfusion}  & (${R_{3D}}$$\to$R)+C &  37.85   & 40.96  & 68.95   &  49.25   &  70.21  & 45.86  & \textbf{89.48}  & 68.52 & 7.1 \\
LoGoNet$^\ddagger$$^*$ \cite{logonet}  & (L$\to$R)+C & 51.72 & 49.00 & 75.54 & 58.75 & 81.19 & 59.68 & 86.15 & 75.67 & 5.8 \\ \hline
RCFusion \cite{rcfusion}  & R+C & 41.70 & 38.95 & 68.31 & 49.65 & 71.87 & 47.50 & 88.33 & 69.23 & $\backslash$ \\
RCBEVDet \cite{rcbevdet}  & R+C & 40.63 & 38.86 & 70.48 & 49.99 & 72.48 & 49.89 & 87.01 & 69.80 & $\backslash$ \\
LXL \cite{lxl} & R+C & 42.33 & 49.48 & 77.12 & 56.31 & 72.18 & 58.30 & 88.31 & 72.93 & 6.1 \\ \hline
CVFusion (Stage1-only) & R+C & 52.53 & 50.76 & 75.80 & 59.70 & 81.30 & 61.44 & 87.23 & 76.66  &6.9 \\
\rowcolor{gray!20} \textbf{CVFusion}$^\ddagger$ \textbf{(Ours)} & R+C & \textbf{60.87} & \textbf{57.89} & \textbf{77.46} & \textbf{65.41} & \textbf{89.86} & \textbf{68.79} & 88.62 & \textbf{82.42} & 5.4\\ \hline
\end{tabular}}
\caption{Experimental results on the VoD \emph{val} set, where ${R_{3D}}$ denotes 3D radar, R denotes 4D radar, C denotes camera and L denotes LiDAR. $\to$ represents the replacement of modality input. ``$\backslash$" means not available. 
$\ddagger$ denotes two-stage networks and methods marked with $*$ are retrained from scratch before testing. 
The best result of each metric is in \textbf{bold} except for PointPillars-L.}
\label{VoD sota comparison}
\end{table*}

\begin{table*}[t]
\centering
\resizebox{0.98\textwidth}{!}{
\begin{tabular}{cc|cccc|c|cccc|c|c}
\hline
    \multirow{2}{*}{Method} & \multirow{2}{*}{Modality} & \multicolumn{5}{c|}{3D (\%)} & \multicolumn{5}{c|}{BEV (\%)} & \multirow{2}{*}{FPS} \\ \cline{3-12} 
 &  & Car & Ped & Cyc & Tru & mAP & Car & Ped & Cyc & Tru & mAP \\ \hline
PointPillars \cite{lang2019pointpillars} & L & 52.67 &  49.07 & 48.59 & 22.74 & 43.27 & 52.76 & 49.23 & 48.32 & 24.72 & 43.76 &$\backslash$\\ \hline
PointPillars$^*$ \cite{lang2019pointpillars}  & L$\to$R & 21.26 &  28.33 & 52.47 & 11.18 & 28.31 & 38.34 & 32.26 & 56.11 & 18.19 & 36.23 &42.9\\ 
SECOND$^*$ \cite{second} & L$\to$R & 18.18 & 24.43 & 32.36 & 14.76 & 22.43 & 36.02 & 28.58 & 39.75 & 19.35 & 30.93 &24.5\\
RPFA-Net \cite{rpfa} & R & 26.89 & 27.36 & 50.95 & 14.46 & 29.91 & 42.89 & 29.81 & 57.09 & 25.98 & 38.94 &$\backslash$\\
SMURF \cite{liu2023smurf} & R & 28.47 & 26.22 & 54.61 & 22.64 & 32.99 & 43.13 & 29.19 & 58.81 & 32.80 & 40.98 &23.1\\
\hline
RCFusion \cite{rcfusion} & R+C & 29.72 &  27.17 & \textbf{54.93} & 23.56 & 33.85 & 40.89 & 30.95 & \textbf{58.30} & 28.92 & 39.76 & 4.7\\ 
LXL \cite{lxl} & R+C & $\backslash$ & $\backslash$ & $\backslash$ & $\backslash$ & 36.32 & $\backslash$  & $\backslash$ & $\backslash$ & $\backslash$ & 41.20 & $\backslash$ \\ \hline
CVFusion (Stage1-only) & R+C & \textbf{54.08} & 26.90 & 40.88 & 22.16 & 36.00 & \textbf{60.22} & 28.17 & 42.82 & 30.96 & 40.54 & 8.7 \\ 
\rowcolor{gray!20} \textbf{CVFusion} \textbf{(Ours)} & R+C & 51.54 & \textbf{29.49} & 49.41 & \textbf{29.55} & \textbf{40.00} & 58.07 & \textbf{31.65} & 51.29 & \textbf{35.29} & \textbf{44.07} & 5.7 \\ \hline
\end{tabular}}
\caption{Evaluation results on the \emph{test} set of TJ4DRadSet, where R denotes 4D imaging radar, C denotes camera and L denotes LiDAR. LXL does not report specific APs for each class. $\to$ represents the replacement of modality input. ``$\backslash$" means not available. 
The methods marked with $*$ are retrained from scratch before testing. 
The best result of each metric is in \textbf{bold} except for PointPillars-L.}
\label{TJ sota comparison}
\end{table*}

\subsubsection{Datasets and Evaluation Metrics}

We conduct experiments on the public View-of-Delft (VoD) \cite{palffy2022multi} and TJ4DRadSet \cite{9922539} dataset. The VoD and TJ4DRadSet are both multi-sensor datasets consisting of LiDAR, camera and 4D radar data recorded in real-world scenes. In the experiments, we follow the official train/val/test set partition. Specifically, the VoD dataset contains 5139, 1296 and 2247 frames for training, validation and testing respectively. Since the official test server for the VoD dataset is not yet released, evaluations are performed on the validation set as usual. In the case of the TJ4DRadSet, the partition is 5717 frames for training and 2040 frames for testing respectively.

The VoD has two types of evaluation metrics. One is the mean Average Precision (mAP) in the entire annotation area (camera FoV up to 50 meters), the other is mAP in the driving corridor area which is defined as \{-4m $<$ $x$ $<$ 4m, 0 $<$ $z$ $<$ 25m\} in camera coordinates. 
In the case of the TJ4DRadSet dataset, evaluation metrics include $\mathrm{mAP_{3D}}$ and $\mathrm{mAP_{BEV}}$ within a range of 70 meters. 
For both datasets, the IoU thresholds for mAP are set to 0.5 for cars and trucks, 0.25 for pedestrians and cyclists .

\subsubsection{Implementation Settings}

For the VoD dataset, the detection range is set [0, 51.2m] on the X axis, [-25.6m, 25.6m] on the Y axis and [-3.0m, 2.0m] on the Z axis. As for the TJ4DRadSet, the detection range is larger, with [0, 70.4m] on the X axis, [-40.0m, 40.0m] on the Y axis and [-4.0m, 2.0m] on the Z axis. 
Voxel size for the radar point cloud is set 0.05m × 0.05m × 0.1m, and the resolution of BEV fusion map before RPN in stage1 is 0.4m × 0.4m.
RoI-pooling in the Stage2 uses a grid number setting of $U = 6$.

\subsubsection{Training and Inference Details}

The implementation is based on OpenPCDet \cite{openpcdet2020} framework, and the model is trained at a batch size of 2, with learning rate 0.01 for 80 epochs on four GTX 3090 GPUs. Except for the frozen Swin-Tiny \cite{liu2021Swin} based 2D backbone that is pretrained on ImageNet-1k, the model is randomly initialized and trained end-to-end from scratch. For data augmentation strategy, we apply random flipping along the X axis, global scaling, and global rotation around the Z axis according to the official recommendations of the OpenPCDet. We follow the setting of \cite{pvrcnn} to determine the effective proposals for the second stage. For post-processing, we use confidence threshold of 0.1 and non-maximum-suppression (NMS) threshold of 0.01.

\begin{table*}[t]
\small
\centering
\resizebox{0.9\textwidth}{!}{
\begin{tabular}{ccccc|ccc|c|ccc|c}
\hline
\multirow{2}{*}{\textbf{Cam}} & \multirow{2}{*}{\textbf{Rad}} & \multirow{2}{*}{\textbf{RGIter}} & \multirow{2}{*}{\textbf{PGF}} & \multirow{2}{*}{\textbf{GGF}}& \multicolumn{4}{c|}{Entire Annotated Area (\%)} & \multicolumn{4}{c}{Driving Corridor Area (\%)} \\ \cline{6-13} 
 & & & & & Car & Ped. & Cyclist & mAP & Car & Ped. & Cyclist & mAP \\ \hline
\checkmark & & & &  & 19.4 & 12.2 & 24.9 & 18.8 & 48.2 &16.4 &41.3 &35.3 \\ 
& \checkmark & & &  & 39.1 & 37.7 & 63.3 & 46.7 & 72.0 & 48.5 & 82.7 & 67.8 \\ 
\checkmark & \checkmark & &  &  & 52.2  & 49.5  & 69.7  & 57.1 &  81.1 & 60.6 & 87.1 & 76.3\\ 
\checkmark & \checkmark & \checkmark &  &   & 52.5  & 50.8  & 75.8  & 59.7 & 81.3  &61.4  &87.2  &76.7  \\ 
\checkmark & \checkmark & \checkmark & \checkmark &  & 56.9 & 54.4 & 73.6 & 61.7 & 89.7 & 66.4 & 88.3 & 81.5 \\ 
\checkmark & \checkmark & \checkmark &  &\checkmark  & 56.3 & 55.2 & 77.4 & 63.0 & 89.5 & 67.8 & 88.4 & 81.9 \\ 
\rowcolor{gray!20} \checkmark & \checkmark & \checkmark & \checkmark & \checkmark & \textbf{60.9} & \textbf{57.9} & \textbf{77.5} & \textbf{65.4} & \textbf{89.9} & \textbf{68.8} & \textbf{88.6} & \textbf{82.4} \\ \hline
\end{tabular}}
\caption{Ablation of the input and fusion modules on the VoD \textit{val} set. \textbf{Cam}, \textbf{Rad}, \textbf{RGIter}, \textbf{PGF} and \textbf{GGF} refer to the camera input, radar input, radar-guided iterative fusion, point-guided fusion and grid-guided fusion, respectively.}
\label{Ablation studies on VoD}
\end{table*}

\begin{table}[t]
\small
\centering
\begin{tabular}{ccc|ccc|c|c}
\hline
\multirow{2}{*}{\textbf{RG}} & \multirow{2}{*}{\textbf{MS}} & \multirow{2}{*}{\textbf{Iter}} & \multicolumn{5}{c}{Entire Annotated Area (\%)} \\ \cline{4-8} &  &  & Car & Ped. & Cyc. & mAP& recall \\ \hline
  &   &  &52.2  &49.5  &69.7  &57.1&72.0  \\ 
\checkmark &   &   &52.3  &49.2  &71.6 &57.7&72.6  \\ 
\checkmark  & \checkmark  & &\textbf{52.6}  &50.4  &73.3  &58.8 &74.3  \\ 
\rowcolor{gray!20} \checkmark  & \checkmark & \checkmark & 52.5 & \textbf{50.8} & \textbf{75.8}  & \textbf{59.7}&\textbf{74.9}  \\ \hline
\end{tabular}
\caption{Ablation of the RGIter module on VoD \textit{val} set. \textbf{RG} and \textbf{MS} denote radar guided and multi-scale but not iterative BEV fusion, respectively. \textbf{Iter} represents iterative fusion mechanism. The recall of proposal boxes is obtained at 3D IoU@0.25.}
\label{tab:RGIter ablation}
\end{table}

\subsection{Comparison with Other Methods}
Currently there are only a few 4D radar-camera fusion methods for comparison. Since LiDAR and radar have similar form of point cloud measurements, the networks designed for LiDAR or 3D radar can be used for 4D radar. We replace the LiDAR or 3D radar input with the 4D radar and retrain the networks to produce experimental results for more comprehensive comparison.

\subsubsection{VoD dataset}
As shown in Table \ref{VoD sota comparison}, our method has made significant improvements on all metrics and achieved new state-of-the-art, far exceeding the previous best 4D radar based method LXL by 9.10\% and 9.49\% mAP in the entire annotated and driving corridor area, respectively. When comparing with the methods originally designed for 3D radar-camera fusion \cite{futr3d, liu2023bevfusion}, our network well utilizes the extra height information of 4D radar and further strengthens them by cross-view fusion. Compared with the two-stage network LoGoNet \cite{logonet} which generates proposal boxes from point-only BEV features, our method also performs much better thanks to the RGIter BEV fusion and elegant cross-view fusion.

\subsubsection{TJ4DRadSet}
Table \ref{TJ sota comparison} shows the evaluation results on the TJ4DRadSet test set. Again, our method outperforms LXL by 3.68\% and 2.87\% in 3D mAP and BEV mAP, respectively, achieving a new state-of-the-art. It fully exhibits the superiority of our cross-view based two–stage fusion mechanism.

As an interesting experiment, we compared our method with the classical LiDAR-based 3D object detection network PointPillars-L on both VoD and TJ4DRadSet. As shown in Table \ref{VoD sota comparison} and \ref{TJ sota comparison}, we are excited to see that our CVFusion performs better than the LiDAR-input PointPillars on most of the metrics, especially in VoD dataset. To the best of our knowledge, it is the first time that a radar-camera fusion model can achieve performance comparable to that of a classical LiDAR-input network.

The inference speed (FPS) is also listed in the last column of the Table \ref{VoD sota comparison} and \ref{TJ sota comparison}. It is worth noting that our stage1-only method infers faster than LXL on the VoD. It should thank to the effective design of the RGIter BEV fusion rather than slower 3D convolution fusion.

\begin{table}[t]
\small
\centering
\begin{tabular}{c|ccc|c}
\hline
\multirow{2}{*}{Method} & \multicolumn{4}{c}{Entire Annotated Area (\%)} \\ \cline{2-5} & Car & Ped. & Cyc. & mAP \\ \hline
Stage1 & 52.5 & 50.8 & \textbf{75.8}  & 59.7   \\ \hline
+ radar features & 52.0 & 50.5 & 73.7 & 58.7   \\ \hline
+ KDE &  52.8 &  50.2 &  74.7 & 59.2  \\ \hline
\rowcolor{gray!20} + CMDA & \textbf{56.9}  & \textbf{54.5} & 73.6  & \textbf{61.7}  \\ \hline
\end{tabular}
\caption{Ablation of the PGF module on VoD \textit{val} set.}
\label{VoD point-guided Fusion}
\end{table}

\subsection{Ablation Studies}
We carried out ablation study on the VoD \textit{val} set. As shown in Table \ref{Ablation studies on VoD}, the performance of the 4D radar-only network is much higher than the camera-only network, showing the importance of the 3D point cloud provided by the radar. Fusing both the camera and the radar by simple BEV concatenation, the performance is greatly improved compared to the single modal network. Integrating the RGIter module can improve mAP by 2.6\% upon the simple concatenation. 
Integration the full stage2 including the PGF and GGF modules can boost mAP by 5.7\% upon the stage1, validating the effectiveness of the two-stage cross-modality fusion pipeline. More detailed ablation within the RGIter, PGF and GGF modules will be shown in the following.

\subsubsection{Ablation of the RGIter}
As shown in Table \ref{tab:RGIter ablation}, compared with direct concatenation of camera and radar BEV features, gradually adding radar guided (\textbf{RG}), multi-scale fusion (\textbf{MS}), and iterative fusion mechanism (\textbf{Iter}), the mAP are increased by 0.6\%, 1.1\%, and 0.9\% respectively. Radar guidance injects the point occupancy information into the image BEV features. Multi-scale fusion further enhances the detection performance of small objects such as cyclists and pedestrians. The iterative mechanism can generate refined multi-scale image BEV features guided by radar to benefit BEV fusion. The recall for the 1st-stage without the RGIter is 72.0\%, and it increases to 74.9\% with RGIter.
The higher mAP and recall of the proposal boxes shows the effectiveness of the RGIter fusion module.

\begin{figure*}[t]
\centering
\includegraphics[width=0.86\textwidth]{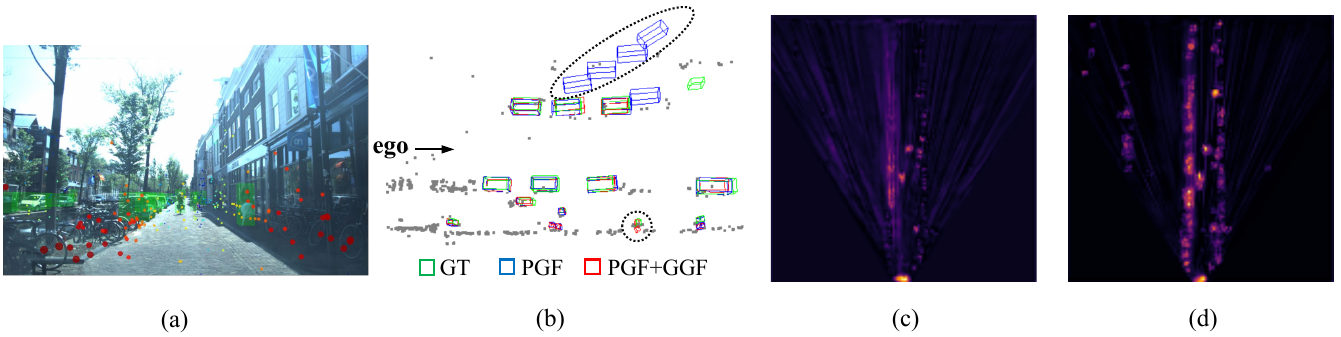}
\caption{Qualitative comparison of the PGF-only and the PGF+GGF (full stage2) detector. The scene image embedded with radar points (a) and its BEV detection results (b), where the significantly different area are marked by the circles. The corresponding BEV feature of the PGF-only (c) and the full stage2 (d). The brighter color corresponding the higher feature value. Best viewed in color.}
\label{+GGF}
\end{figure*}

\subsubsection{Ablation on PGF}
As shown in Table \ref{VoD point-guided Fusion}, the performance is degraded upon the stage1 results if the PGF merely use radar features. 
It can be attributed to the high sparsity and noises associated with the radar point measurements. After adding KDE features, performance can be improved by 0.5\% mAP, but is still lower than the stage1. When CMDA is introduced, the performance is greatly improved by 2.5\% mAP, thanks to the rich semantic information induced by the images.

\begin{table}[t]
\centering
\small
\begin{tabular}{ccc|ccc|c}
\hline
\multirow{2}{*}{\textbf{CD1}} & \multirow{2}{*}{\textbf{CD2}} & \multirow{2}{*}{\textbf{SA}} & \multicolumn{4}{c}{Entire Annotated Area (\%)} \\ \cline{4-7} &  &  & Car & Ped. & Cyc. & mAP \\ \hline
  &   &  & 56.9 & 54.5 & 73.6 & 61.7 \\ 
\checkmark &   &  & 54.0 & 57.2 & 77.1 & 62.8 \\ 
\checkmark  & \checkmark  & & 59.4 & 57.1 & 76.6 & 64.4  \\ 
\rowcolor{gray!20} \checkmark  & \checkmark & \checkmark & \textbf{60.9} & \textbf{57.9} & \textbf{77.5}  & \textbf{65.4}  \\ \hline
\end{tabular}
\caption{Ablation of the GGF module on VoD \textit{val} set. \textbf{CD1} and \textbf{CD2} denote the first and second CMDA block, respectively. \textbf{SA} represents Self-Attention.}
\label{vod GGF}
\end{table}
\subsubsection{Ablation on GGF}

The ablations of GGF module is shown in Table \ref{vod GGF}. Fusing only the image feature to the grids with CMDA block gives 1.1\% mAP improvement to that of PGF, which preliminary shows the necessity of the grid-guided fusion. Further fusing the BEV features by the second CMDA block produces more improvements. This should thank to the valuable information and constraints provided by the BEV features from the stage1. Finally, the self attention block helps interact the features generated by the two fusion modules, resulting in a total performance gain of 3.7\% mAP over the stage1 + PGF baseline.

A qualitative comparison between the PGF-only and the full stage2 detector is shown in \cref{+GGF}. It can be seen in \cref{+GGF} (a) and (b) that the PGF-only cannot well handle the proposals with few radar points and almost retains them without any refinement. After introducing the GGF module, a large number of false positive boxes are depressed. Meanwhile, more accurate regressions are produced thanks to the finely integration of cross-view features. \cref{+GGF} (c) and (d) shows the visualization effect of the feature map before and after adding the GGF module.

\subsubsection{Ablation on Image Backbones}
\begin{table}[t]
\small
\centering
\begin{tabular}{c|c|c|c}
\hline
\textbf{Methods} &\textbf{Backbones} & \textbf{mAP}&\textbf{Ours} \\ \hline
RCFusion\cite{rcfusion} &\multirow{2}{*}{ResNet-50\cite{resnet}} &49.65&\multirow{2}{*}{\textbf{60.29}}  \\
RCBEVDet\cite{rcbevdet} & &49.99&  \\ \hline
FUTR3D\cite{futr3d} &ResNet-101\cite{resnet} &49.03&\textbf{62.96}  \\ \hline
BEVFusion\cite{liu2023bevfusion} &\multirow{2}{*}{Swin-Tiny\cite{liu2021Swin}} &49.25&\multirow{2}{*}{\textbf{65.41}} \\ 
LoGoNet\cite{logonet} & &58.78& \\ \hline
\end{tabular}
\caption{Ablations of the image backbones on VoD \textit{val} set.}
\label{tab:Backbones ablations}
\end{table}

It is possible that different 2D image backbone can influence the overall performance of the network. To further independently verify the contribution of our design, we carry out the ablation on image backbones and compare with the existing methods. We use the available open source code of the state-of-the-art methods in Table \ref{VoD sota comparison} and replace our Swin-Tiny \cite{liu2021Swin} backbone with theirs.  The results are shown in Table \ref{tab:Backbones ablations}.  As expected, the backbone does play a important role in the final performance. However, regardless of whether the ResNet-50 or ResNet-101 backbones is integrated, our design still greatly outperforms the competitors. 

\section{Conclusion}
We propose a novel two-stage-based cross-view fusion network for 3D object detection with 4D radar and camera. Radar Guided Iterative BEV fusion provides high-quality proposal boxes, and the PGF- and the GGF-based stage2 fusion leverages various forms of features including points, FV and BEV, providing comprehensive clues for the refinement of the proposals. Experimental results show that our network outperforms the state-of-the-art methods by a large margin and even achieves comparable performance with LiDAR-input method, revealing the enormous potential of 4D radar and camera fusion in autonomous driving.

{
    \small
    \bibliographystyle{ieeenat_fullname}
    \bibliography{main}
}

\end{document}